\documentclass[11pt,a4paper]{article}
\usepackage[hyperref]{acl2019}
\usepackage{times}
\usepackage{latexsym}
\usepackage{multirow}
\usepackage{graphicx}
\usepackage{subfigure}
\usepackage{enumitem}
\usepackage{color}
\usepackage{bm}
\usepackage{amsmath}
\usepackage{amsfonts}
\usepackage{array}
\usepackage{url}
\usepackage{amsthm}
\usepackage{amssymb}

\theoremstyle{definition}

\usepackage{etoolbox}

\aclfinalcopy 

\title{Recent Advances in Neural Question Generation}

\author{
Liangming Pan, Wenqiang Lei, Tat-Seng Chua and Min-Yen Kan \\
School of Computing \\
National University of Singapore \\
Singapore 117417 \\
{\tt \{e0272310,wenql,kanmy,chuats\}comp.nus.edu.sg} \\
}

\date{}

\begin{document}
\maketitle
\begin{abstract}

Emerging research in Neural Question Generation (NQG) has started to integrate a larger variety of inputs, and generating questions requiring higher levels of cognition. These trends point to NQG as a bellwether for NLP, about how human intelligence embodies the skills of curiosity and integration. 

We present a comprehensive survey of neural question generation, examining the corpora, methodologies, and evaluation methods. From this, we elaborate on what we see as emerging on NQG's trend: in terms of the learning paradigms, input modalities, and cognitive levels considered by NQG.  We end by pointing out the potential directions ahead. 

\end{abstract}

\section{Introduction}
\label{sec:introduction}

Question Generation (QG) concerns the task of ``automatically generating questions from various inputs such as raw text, database, or semantic representation"~\cite{rus2008question}. 
People have the ability to ask rich, creative, and revealing questions~\cite{DBLP:conf/nips/RotheLG17}; \textit{e.g.}, asking \textit{Why did Gollum betray his master Frodo Baggins?} after reading the fantasy novel \textit{The Lord of the Rings}.  How can machines be endowed with the ability to ask relevant and to-the-point questions, given various inputs?
This is a challenging, complementary task to Question Answering (QA). Both QA and QG require an in-depth understanding of the input source and the ability to reason over relevant contexts. But beyond understanding, QG additionally integrates the challenges of Natural Language Generation (NLG), \textit{i.e.}, generating grammatically and semantically correct questions. 

QG is of practical importance: in education, forming good questions are crucial for evaluating students’ knowledge and stimulating self-learning. QG can generate assessments for course materials~\cite{DBLP:conf/naacl/HeilmanS10} or be used as a component in adaptive, intelligent tutoring systems~\cite{DBLP:conf/enlg/LindbergPNW13}.  In dialog systems, fluent QG is an important skill for chatbots, \textit{e.g.}, in initiating conversations or obtaining specific information from human users. 
QA and reading comprehension also benefit from QG, by reducing the needed human labor for creating large-scale datasets.  
We can say that traditional QG mainly focused on generating factoid questions from a single sentence or a paragraph, spurred by a series of workshops during 2008--2012~\cite{DBLP:conf/aied/RusL09,rus2010overview,DBLP:conf/enlg/RusWPLSM11,DBLP:journals/dad/RusWPLSM12}. 

Recently, driven by advances in deep learning, QG research has also begun to utilize ``neural'' techniques, to develop end-to-end neural models to generate deeper questions~\cite{DBLP:conf/icwsm/ChenYHH18} and to pursue broader applications~\cite{DBLP:conf/acl/SerbanGGACCB16,DBLP:conf/acl/MostafazadehMDM16}. 

While there have been considerable advances made in NQG, the area lacks a comprehensive survey. 
This paper fills this gap by presenting a systematic survey on recent development of NQG, focusing on three emergent trends that deep learning has brought in QG: (1) the change of learning paradigm, (2) the broadening of the input spectrum, and (3) the generation of deep questions. 

\section{Fundamental Aspects of NQG}
\label{sec:overview}
\vspace{-0.2cm}

For the sake of clean exposition, we first provide a broad overview of QG by conceptualizing the problem from the perspective of the three introduced aspects: (1) its learning paradigm, (2) its input modalities, and (3) the cognitive level it involves. This combines past research with recent trends, providing insights on how NQG connects to traditional QG research. 

\subsection{Learning Paradigm}
\label{sec:focus_of_QG}

QG research traditionally considers two fundamental aspects in question asking: ``What to ask'' and ``How to ask''. A typical QG task considers the identification of the important aspects to ask about (``what to ask''), and learning to realize such identified aspects as natural language (``how to ask''). Deciding what to ask is a form of machine understanding: a machine needs to capture important information dependent on the target application, akin to automatic summarization. Learning how to ask, however, focuses on aspects of the language quality such as grammatical correctness, semantically preciseness and language flexibility.

Past research took a reductionist approach, separately considering these two problems of ``what'' and ``how'' via \textit{content selection} and \textit{question construction}. 
Given a sentence or a paragraph as input, content selection selects a particular salient topic worthwhile to ask about and determines the question type (\textit{What}, \textit{When}, \textit{Who}, etc.).  Approaches either take a  syntactic~\cite{gates2008generating,DBLP:conf/its/LiuCR10,heilman2011automatic} 
or semantic~\cite{DBLP:journals/dad/YaoBZ12,DBLP:conf/enlg/LindbergPNW13,DBLP:conf/acl/MazidiN14,DBLP:journals/coling/ChaliH15}
tack, both starting by applying syntactic or semantic parsing, respectively, to obtain intermediate symbolic representations. Question construction then converts intermediate representations to a natural language question, taking either a {\it tranformation-} or {\it template-based} approach.
The former~\cite{ali2010automation,pal2010qgstec,heilman2011automatic} rearranges the surface form of the input sentence to produce the question; the latter~\cite{chen2009aist,DBLP:journals/dad/LiuCR12,DBLP:conf/acl/RokhlenkoS13} generates questions from pre-defined question templates. Unfortunately, such QG architectures are limiting, as their representation is confined to the variety of intermediate representations, transformation rules or templates.  

In contrast, neural models motivate an end-to-end architectures.  Deep learned frameworks contrast with the reductionist approach, admitting approaches that jointly optimize for both the ``what'' and ``how'' in an unified framework. The majority of current NQG models follow the sequence-to-sequence (Seq2Seq) framework that use a unified representation and joint learning of content selection (via the encoder) and question construction (via the decoder). In this framework, traditional parsing-based content selection has been replaced by more flexible approaches such as attention~\cite{DBLP:journals/corr/BahdanauCB14} and copying mechanism~\cite{DBLP:conf/acl/GulcehreANZB16}. Question construction has become completely data-driven, requiring far less labor compared to transformation rules, enabling better language flexibility compared to question templates. 

However, unlike other Seq2Seq learning NLG tasks, such as Machine Translation, Image Captioning, and Abstractive Summarization, which can be loosely regarded as learning a one-to-one mapping, generated questions can differ significantly when the intent of asking differs (\textit{e.g.}, the target answer, the target aspect to ask about, and the question's depth). 
In Section~\ref{sec:Method}, we summarize different NQG methodologies based on Seq2Seq framework, investigating how some of these QG-specific factors are integrated with neural models, and discussing what could be further explored. 
The change of learning paradigm in NQG era is also represented by multi-task learning with other NLP tasks, for which we discuss in Section~\ref{sec:multi_task}. 

\subsection{Input Modality}

Question generation is an NLG task for which the input has a wealth of possibilities depending on applications. 
While a host of input modalities have been considered in other NLG tasks, such as text summarization~\cite{mani1999advances}, image captioning~\cite{DBLP:conf/cvpr/VinyalsTBE15} and table-to-text generation~\cite{DBLP:conf/emnlp/LebretGA16}, traditional QG mainly focused on textual inputs, especially declarative sentences, explained by the original application domains of question answering and education, which also typically featured textual inputs. 

Recently, with the growth of various QA applications such as Knowledge Base Question Answering (KBQA)~\cite{DBLP:journals/pvldb/CuiXWSHW17} and Visual Question Answering (VQA)~\cite{DBLP:conf/iccv/AntolALMBZP15}, 
NQG research has also widened the spectrum of sources to include knowledge bases~\cite{DBLP:conf/eacl/KhapraRJR17} and images~\cite{DBLP:conf/acl/MostafazadehMDM16}.
This trend is also spurred by the remarkable success of neural models in feature representation, especially on image features~\cite{DBLP:conf/nips/KrizhevskySH12} and knowledge representations~\cite{DBLP:conf/nips/BordesUGWY13}. 
We discuss adapting NQG models to other input modalities in Section~\ref{sec:input_modalities}. 

\subsection{Cognitive Levels}
\label{sec:Cognitive_Level}
Finally, we consider the required cognitive process behind question asking, a distinguishing factor for questions~\cite{anderson2001taxonomy}. 
A typical framework that attempts to categorize the cognitive levels involved in question asking comes from Bloom's taxonomy~\cite{bloom1984taxonomy}, which has undergone several revisions and currently has six cognitive levels: {\it Remembering}, {\it Understanding}, {\it Applying}, {\it Analyzing}, {\it Evaluating} and {\it Creating}~\cite{anderson2001taxonomy}. 

Traditional QG focuses on shallow levels of Bloom's taxonomy: typical QG research is on generating sentence-based factoid questions (\textit{e.g.}, \textit{Who}, \textit{What}, \textit{Where} questions), whose answers are simple constituents in the input sentence~\cite{DBLP:conf/naacl/HeilmanS10,heilman2011automatic}. 
However, a QG system achieving human cognitive level should be able to generate meaningful questions that cater to higher levels of Bloom's taxonomy~\cite{DBLP:conf/acl-tea/DesaiDM18}, such as \textit{Why}, \textit{What-if}, and \textit{How} questions. Traditionally, those ``deep'' questions are generated through shallow methods such as handcrafted templates~\cite{DBLP:journals/dad/LiuCR12,DBLP:conf/acl/RokhlenkoS13}; however, these methods lack a real understanding and reasoning over the input. 

Although asking deep questions is complex, NQG's ability to generalize over voluminous data has enabled recent research to explore the comprehension and reasoning aspects of QG~\cite{DBLP:conf/acl/LabutovBV15,DBLP:conf/nips/RotheLG17,DBLP:conf/icwsm/ChenYHH18,DBLP:conf/acl-tea/DesaiDM18}. We investigate this trend in Section~\ref{sec:Limitations}, examining the limitations of current Seq2Seq model in generating deep questions, and the efforts made by existing works, indicating further directions ahead. 

The rest of this paper provides a systematic survey of NQG, covering corpus and evaluation metrics before examining specific neural models. 

\begin{table*}
    \small
    \renewcommand\arraystretch{1.1}
	\begin{center}
		\begin{tabular}{|c|c|c|c|c|c|c|} \hline
		    Cognitive &
			\multirow {2}{*}{Dataset / Contributor} &
            Answer &
			\multirow {2}{*}{Domain} &
			\multicolumn{3}{c|}{Statistics} \\ \cline{5-7} 
		    Level & & Type & & Documents & Questions & Q./Doc \\ \hline \hline
           \multirow {2}{*}{Shallow} & SQuAD~\cite{DBLP:conf/emnlp/RajpurkarZLL16} & text span & Wikipedia & 20,958 & 97,888 & 4.67 \\ \cline{2-7}
           & NewsQA~\cite{DBLP:conf/rep4nlp/TrischlerWYHSBS17} & text span & News & 12,744 & 119,633 & 9.39 \\ \hline
           \multirow {2}{*}{Medium} & MS MARCO~\cite{DBLP:conf/nips/NguyenRSGTMD16} & human generated & Web article & 1,010,916 & 3,563,535 & 3.53 \\ \cline{2-7}
           & RACE~\cite{DBLP:conf/emnlp/LaiXLYH17} & multiple choice & Education & 27,933 & 72,547 & 2.60 \\ \hline
           \multirow {2}{*}{Deep} & LearningQ~\cite{DBLP:conf/icwsm/ChenYHH18} & no answer & Education & 10,841 & 231,470 & 21.35 \\ \cline{2-7}
           & NarrativeQA~\cite{DBLP:journals/tacl/KociskySBDHMG18} & human generated & Story & 1,572 & 46,765 & 29.75 \\ \hline
		\end{tabular}
	\end{center}
	\vspace{-0.2cm}
	\caption{NQG datasets grouped by their cognitive level and answer type, where the number of documents, the number of questions, and the average number of questions per document (Q./Doc) for each corpus are listed. }
	\label{tbl:NQG_Datasets}
	\vspace{-0.4cm}
\end{table*}

\section{Corpora}
\vspace{-0.15cm}

As QG can be regarded as a dual task of QA, in principle any QA dataset can be used for QG as well. However, there are at least two corpus-related factors that affect the difficulty of question generation. The first is the required \textbf{cognitive level} to answer the question, as we discussed in the previous section. Current NQG has achieved promising results on datasets consisting mainly of shallow factoid questions, such as SQuAD~\cite{DBLP:conf/emnlp/RajpurkarZLL16} and MS MARCO~\cite{DBLP:conf/nips/NguyenRSGTMD16}. However, the performance drops significantly on deep question datasets, such as LearningQ~\cite{DBLP:conf/icwsm/ChenYHH18}, shown in Section~\ref{sec:Limitations}. The second factor is the \textbf{answer type}, i.e., the expected form of the answer, typically having four settings: (1) the answer is a text span in the passage, which is usually the case for factoid questions, (2) human-generated, abstractive answer that may not appear in the passage, usually the case for deep questions, (3) multiple choice question where question and its distractors should be jointly generated, and (4) no given answer, which requires the model to automatically learn what is worthy to ask. The design of NQG system differs accordingly.  

Table~\ref{tbl:NQG_Datasets} presents a listing of the NQG corpora grouped by their cognitive level and answer type, along with their statistics. 
Among them, SQuAD was used by most groups as the benchmark to evaluate their NQG models. This provides a fair comparison between different techniques. However, it raises the issue that most NQG models work on factoid questions with answer as text span, leaving other types of QG problems less investigated, such as generating deep multi-choice questions. To overcome this, a wider variety of corpora should be benchmarked against in future NQG research. 

\section{Evaluation Metrics}
\vspace{-0.1cm}

Although the datasets are commonly shared between QG and QA, it is not the case for evaluation: it is challenging to define a gold standard of proper questions to ask. Meaningful, syntactically correct, semantically sound and natural are all useful criteria, yet they are hard to quantify. Most QG systems involve \textit{human evaluation}, commonly by randomly sampling a few hundred generated questions, and asking human annotators to rate them on a $5$-point Likert scale. The average rank or the percentage of best-ranked questions are reported and used for quality marks. 

As human evaluation is time-consuming, common automatic evaluation metrics for NLG, such as BLEU~\cite{DBLP:conf/acl/PapineniRWZ02}, METEOR~\cite{DBLP:journals/mt/LavieD09}, and ROUGE~\cite{lin2004rouge}, are also widely used. However, some studies~\cite{DBLP:conf/eacl/Callison-BurchOK06,DBLP:conf/emnlp/LiuLSNCP16} have shown that these metrics do not correlate well with fluency, adequacy, coherence, as they essentially compute the $n$-gram similarity between the source sentence and the generated question. To overcome this, \citet{DBLP:conf/emnlp/NemaK18} proposed a new metric to evaluate the ``answerability'' of a question by calculating the scores for several question-specific factors, including question type, content words, function words, and named entities. However, as it is newly proposed, it has not been applied to evaluate any NQG system yet. 

To accurately measure what makes a good question, especially deep questions, improved evaluation schemes are required to specifically investigate the mechanism of question asking. 

\section{Methodology}
\label{sec:Method}
\vspace{-0.1cm}

Many current NQG models follow the Seq2Seq architecture. 
Under this framework, given a passage (usually a sentence) $X = (x_1, \cdots, x_n)$ and (possibly) a target answer $A$ (a text span in the passage) as input, an NQG model aims to generate a question $Y = (y_1, \cdots, y_m)$ asking about the target answer $A$ in the passage $X$, which is defined as finding the best question $\bar Y$ that maximizes the conditional likelihood given the passage $X$ and the answer $A$: 

\vspace{-0.5cm}
\begin{align}
\label{equ:NQG_problem}
\bar Y & = \arg \max_Y P(Y \vert X, A) \\
\vspace{-0.5cm}
& = \arg \max_Y \sum_{t=1}^m P(y_t \vert X, A, y_{< t})
\end{align}
\vspace{-0.5cm}

\noindent \citet{DBLP:conf/acl/DuSC17} pioneered the first NQG model using an attention Seq2Seq model~\cite{DBLP:journals/corr/BahdanauCB14}, which feeds a sentence into an RNN-based encoder, and generate a question about the sentence through a decoder. The attention mechanism is applied to help decoder pay attention to the most relevant parts of the input sentence while generating a question. Note that this base model does not take the target answer as input. 
Subsequently, neural models have adopted attention mechanism as a default~\cite{DBLP:conf/nlpcc/ZhouYWTBZ17,DBLP:conf/emnlp/DuanTCZ17,DBLP:conf/inlg/HarrisonW18}. 

Although these NQG models all share the Seq2Seq framework, they differ in the consideration of --- (1) QG-specific factors (\textit{e.g.}, answer encoding, question word generation, and paragraph-level contexts), and (2) common NLG techniques (\textit{e.g.}, copying mechanism, linguistic features, and reinforcement learning) --- discussed next. 

\subsection{Encoding Answers}
\label{sec:encode_answer}

The most commonly considered factor by current NQG systems is the target answer, which is typically taken as an additional input to guide the model in deciding which information to focus on when generating; otherwise, the NQG model tend to generate questions without specific target (\textit{e.g.}, ``What is mentioned?"). 
Models have solved this by either treating the answer's position as an extra input feature~\cite{DBLP:conf/nlpcc/ZhouYWTBZ17,DBLP:conf/emnlp/ZhaoNDK18}, or by encoding the answer with a separate RNN~\cite{DBLP:conf/emnlp/DuanTCZ17,DBLP:journals/corr/abs-1809-02393}. 

The first type of method augments each input word vector with an extra \textit{answer indicator feature}, indicating whether this word is within the answer span. ~\citet{DBLP:conf/nlpcc/ZhouYWTBZ17} implement this feature using the BIO tagging scheme, while ~\citet{DBLP:conf/inlg/HarrisonW18} directly use a binary indicator. 
In addition to the target answer, ~\citet{DBLP:conf/emnlp/SunLLHMW18} argued that the context words closer to the answer also deserve more attention from the model, since they are usually more relevant. 
To this end, they incorporate trainable position embeddings $(d_{p_1}, d_{p_2}, \cdots, d_{p_n})$ into the computation of attention distribution, where $p_i$ is the relative distance between the $i$-th word and the answer, and $d_{p_i}$ is the embedding of $p_i$. 
This achieved an extra BLEU-4 gain of $0.89$ on SQuAD. 

To generate answer-related questions, extra answer indicators explicitly emphasize the importance of answer; however, it also increases the tendency that generated questions include words from the answer, resulting in useless questions, as observed by ~\citet{DBLP:journals/corr/abs-1809-02393}. 
For example, given the input ``John Francis O’Hara was elected president of Notre Dame in 1934.", an improperly generated question would be ``Who was elected John Francis?", which exposes some words in the answer.
To address this, they propose to replace the answer into a special token for passage encoding, and a separate RNN is used to encode the answer. 
The outputs from two encoders are concatenated as inputs to the decoder. ~\citet{DBLP:conf/naacl/SongWHZG18} adopted a similar idea that separately encodes passage and answer, but they instead use the multi-perspective matching between two encodings as an extra input to the decoder. 

We forecast treating the passage and the target answer separately as a future trend, as it results in a more flexible model, which generalizes to the abstractive case when the answer is not a text span in the input passage. However, this inevitably increases the model complexity and difficulty in training. 

\vspace{-0.1cm}
\subsection{Question Word Generation}

Question words (\textit{e.g.}, ``when'', ``how'', and ``why'') also play a vital role in QG; ~\citet{DBLP:conf/emnlp/SunLLHMW18} observed that the mismatch between generated question words and answer type is common for current NQG systems. 
For example, a when-question should be triggered for answer ``the end of the Mexican War" while a why-question is generated by the model. 
A few works~\cite{DBLP:conf/emnlp/DuanTCZ17,DBLP:conf/emnlp/SunLLHMW18} considered question word generation separately in model design. 

\citet{DBLP:conf/emnlp/DuanTCZ17} proposed to first generate a question template that contains question word (\textit{e.g.}, ``how to \#", where \# is the placeholder), before generating the rest of the question. To this end, they train two Seq2Seq models; the former learns to generate question templates for a given text 
, while the latter learns to fill the blank of template to form a complete question. Instead of a two-stage framework, ~\citet{DBLP:conf/emnlp/SunLLHMW18} proposed a more flexible model by introducing an additional decoding mode that generates the question word. 
When entering this mode, the decoder produces a question word distribution based on a restricted set of vocabulary using the answer embedding, the decoder state, and the context vector. 
The switch between different modes is controlled by a discrete variable produced by a learnable module of the model in each decoding step. 

Determining the appropriate question word harks back to question type identification, which is correlated with the question intention, as different intents may yield different questions, even when presented with the same (passage, answer) input pair. 
This points to the direction of exploring question pragmatics, where external contextual information (such as intent) can inform and influence how questions should optimally be generated. 

\subsection{Paragraph-level Contexts}

Leveraging rich paragraph-level contexts around the input text is another natural consideration to produce better questions. According to~\cite{DBLP:conf/acl/DuSC17}, around 20\% of questions in SQuAD require paragraph-level information to be answered.
However, as input texts get longer, Seq2Seq models have a tougher time effectively utilizing relevant contexts, while avoiding irrelevant information.

To address this challenge, ~\citet{DBLP:conf/emnlp/ZhaoNDK18} proposed a gated self-attention encoder to refine the encoded context by fusing important information with the context's self-representation properly, which has achieved state-of-the-art results on SQuAD. The long passage consisting of input texts and its context is first embedded via LSTM with answer position as an extra feature. The encoded
representation is then fed through a gated self-matching network~\cite{DBLP:conf/acl/WangYWCZ17} to aggregate information from the entire passage and embed intra-passage dependencies. Finally, a feature fusion gate~\cite{DBLP:conf/acl/GongB18} chooses relevant information between the original and self-matching enhanced representations. 

Instead of leveraging the whole context, ~\citet{DBLP:conf/acl/CardieD18} performed a pre-filtering by running a coreference resolution system on the context passage to obtain coreference clusters for both the input sentence and the answer. The co-referred sentences are then fed into a gating network, from which the outputs serve as extra features to be concatenated with the original input vectors. 

\vspace{-0.1cm}
\subsection{Answer-unaware QG}
\vspace{-0.1cm}

The aforementioned models require the target answer as an input, in which the answer essentially serves as the focus of asking. However, in the case that only the input passage is given, a QG system should automatically identify question-worthy parts within the passage.  This task is synonymous with content selection in traditional QG. 
To date, only two works~\cite{DBLP:conf/emnlp/DuC17,DBLP:conf/acl/SubramanianWYZT18} have worked in this  setting. 
They both follow the traditional decomposition of QG into content selection and question construction but implement each task using neural networks. 
For content selection, ~\citet{DBLP:conf/emnlp/DuC17} learn a sentence selection task to identify question-worthy sentences from the input paragraph using a neural sequence tagging model. ~\citet{DBLP:conf/acl/SubramanianWYZT18} train a neural keyphrase extractor to predict keyphrases of the passage. For question construction, they both employed the Seq2Seq model, for which the input is either the selected sentence or the input passage with keyphrases as target answer. 

However, learning what aspect to ask about is quite challenging when the question requires reasoning over multiple pieces of information within the passage; {\it cf} the Gollum question from the introduction. Beyond retrieving question-worthy information, we believe that studying how different reasoning patterns (e.g., inductive, deductive, causal and analogical) affects the generation process will be an aspect for future study. 

\subsection{Technical Considerations}

Common techniques of NLG have also been considered in NQG model, summarized as $3$ tactics: 

\noindent \textbf{1. Copying Mechanism.} Most NQG models~\cite{DBLP:conf/nlpcc/ZhouYWTBZ17,DBLP:conf/rep4nlp/YuanWGSBZST17,DBLP:conf/lats/WangLNWGB18,DBLP:conf/inlg/HarrisonW18,DBLP:conf/pakdd/KumarBMRL18} employ the \textit{copying mechanism} of ~\citet{DBLP:conf/acl/GulcehreANZB16}, which directly copies relevant words from the source sentence to the question during decoding. This idea is widely accepted as it is common to refer back to phrases and entities appearing in the text when formulating factoid questions, and difficult for a RNN decoder to generate such rare words on its own. 

\noindent \textbf{2. Linguistic Features.} Approaches also seek to leverage additional linguistic features that complements word embeddings, including word case, POS and NER tags~\cite{DBLP:conf/nlpcc/ZhouYWTBZ17,DBLP:conf/lats/WangLNWGB18} as well as coreference~\cite{DBLP:conf/inlg/HarrisonW18} and dependency information~\cite{DBLP:conf/pakdd/KumarBMRL18}. These categorical features are vectorized and concatenated with word embeddings. The feature vectors can be either one-hot or trainable and serve as input to the encoder. 

\noindent \textbf{3. Policy Gradient.}  Optimizing for just ground-truth log likelihood ignores the many equivalent ways of asking a question.  Relevant QG work~\cite{DBLP:conf/rep4nlp/YuanWGSBZST17,DBLP:journals/corr/abs-1808-04961} have adopted policy gradient methods to add task-specific rewards (such as BLEU or ROUGE) to the original objective.  This helps to diversify the questions generated, as the model learns to distribute probability mass among equivalent expressions rather than the single ground truth question. 

\begin{table*}
    \small
	\begin{center}
		\begin{tabular}{ |c|c|c|c|c|c|c|c|c|c|} \hline
			\multirow {2}{*}{Models}  &
			\multirow {2}{*}{Answer Encoding}  &
			\multicolumn{5}{c|}{Features} & \multicolumn{3}{c|}{Performance} \\ 
			\cline{3-10}
		    &  & QW & PC & CP & LF & PG & BLEU-4 & METEOR & ROUGE$_\text{L}$ \\ 
		    \hline \hline
		    Du et al.~\shortcite{DBLP:conf/acl/DuSC17} & not used &  & & & & & 12.28 & 16.62 & 39.75 \\ \hline
		   Duan et al.~\shortcite{DBLP:conf/emnlp/DuanTCZ17} & not used & $\bullet$ &  &  &  &  & 12.28 & $-$ & $-$ \\ \hline
		    Zhou et al.~\shortcite{DBLP:conf/nlpcc/ZhouYWTBZ17} & answer position &  &  & $\bullet$ & $\bullet$ &  & 13.29 & $-$ & $-$ \\ \hline
		    Yuan et al.~\shortcite{DBLP:conf/rep4nlp/YuanWGSBZST17} & answer position & & & $\bullet$ &  & $\bullet$ & 10.50 & $-$ & $-$ \\ \hline
		    Wang et al.~\shortcite{DBLP:conf/lats/WangLNWGB18} & answer position &  &  & $\bullet$ & $\bullet$ &  & 13.86 & 18.38 & 44.37 \\ \hline
		    Harrsion et al.~\shortcite{DBLP:conf/inlg/HarrisonW18} & answer position &  &  & $\bullet$ & $\bullet$ &  & 14.39 & 19.54 & 43.00 \\ \hline
		    Kumar et al.~\shortcite{DBLP:journals/corr/abs-1808-04961} & not used &  &  & $\bullet$ & $\bullet$ & $\bullet$ & 16.17 & 19.85 & 43.90 \\ \hline
		    Sun et al.~\shortcite{DBLP:conf/emnlp/SunLLHMW18} & answer+context position & $\bullet$ &  & $\bullet$ &  &  & 15.64 & $-$ & $-$ \\ \hline
		    Zhao et al.~\shortcite{DBLP:conf/emnlp/ZhaoNDK18} & answer position & & $\bullet$ & $\bullet$ & & & \textbf{16.38} & \textbf{20.25} & \textbf{44.48} \\ \hline
		    Du and Cardie~\shortcite{DBLP:conf/acl/CardieD18} & answer position & & $\bullet$ & $\bullet$ & & & 15.16 & 19.12 & $-$ \\ \hline
		    Song et al.~\shortcite{DBLP:conf/naacl/SongWHZG18} & separate encoder & & & $\bullet$ & & & 13.98 & 18.77 & 42.72 \\ \hline
		    Kim et al.~\shortcite{DBLP:journals/corr/abs-1809-02393} & separate encoder &  &  &  &  &  & 16.20 & 19.92 & 43.96 \\
		    \hline
		\end{tabular}
	\end{center}
	\vspace{-0.1cm}
	\caption{Existing NQG models with their best-reported performance on SQuAD. Legend: \textbf{QW}: question word generation, \textbf{PC}: paragraph-level context, \textbf{CP}: copying mechanism, \textbf{LF}: linguistic features, \textbf{PG}: policy gradient. }
	\label{tbl:NQG_Summary}
	\vspace{-0.2cm}
\end{table*}

\subsection{The State of the Art}

In Table~\ref{tbl:NQG_Summary}, we summarize existing NQG models with their employed techniques and their best-reported performance on SQuAD. These methods achieve comparable results; as of this writing, ~\citet{DBLP:conf/emnlp/ZhaoNDK18} is the state-of-the-art. 

Two points deserve mention. 
First, while the copying mechanism has shown marked improvements, there exist shortcomings.
\citet{DBLP:journals/corr/abs-1809-02393} observed many invalid answer-revealing questions attributed to the use of the copying mechanism; \textit{cf} the John Francis example in Section~\ref{sec:encode_answer}. 
They abandoned copying but still achieved a performance rivaling other systems. In parallel application areas such as machine translation, the copy mechanism has been to a large extent replaced with self-attention~\cite{DBLP:journals/corr/LinFSYXZB17}
or transformer~\cite{DBLP:conf/nips/VaswaniSPUJGKP17}. 
The future prospect of the copying mechanism requires further investigation. 
Second, recent approaches that employ paragraph-level contexts have shown promising results: not only boosting performance, but also constituting a step towards deep question generation, which requires reasoning over rich contexts. 

\vspace{-0.1cm}
\section{Emerging Trends}
\label{sec:Applications}

We discuss three trends that we wish to call practitioners' attention to as NQG evolves to take the center stage in QG: Multi-task Learning, Wider Input Modalities and Deep Question Generation.
\vspace{-0.1cm}
\subsection{Multi-task Learning}
\label{sec:multi_task}

As QG has become more mature, work has started to investigate how QG can assist in other NLP tasks, and vice versa. 
Some NLP tasks benefit from enriching training samples by QG to alleviate the data shortage problem. 
This idea has been successfully applied to semantic parsing~\cite{DBLP:conf/emnlp/GuoSTDYCCCZ18} and QA~\cite{DBLP:conf/naacl/SachanX18}. In the semantic parsing task that maps a natural language question to a SQL query, ~\citet{DBLP:conf/emnlp/GuoSTDYCCCZ18} achieved a 3$\%$ performance gain with an enlarged training set that contains pseudo-labeled $(SQL, question)$ pairs generated by a Seq2Seq QG model. In QA, ~\citet{DBLP:conf/naacl/SachanX18} employed the idea of self-training~\cite{DBLP:conf/cikm/NigamG00} to jointly learn QA and QG. The QA and QG models are first trained on a labeled corpus. Then, the QG model is used to create more questions from an unlabeled text corpus and the QA model is used to answer these newly-created questions. The newly-generated question--answer pairs form an enlarged dataset to iteratively retrain the two models. The process is repeated while performance of both models improve. 

Investigating the core aspect of QG, we say that a well-trained QG system should have the ability to: (1) find the most salient information in the passage to ask questions about, and (2) given this salient information as target answer, to generate an answer related question. ~\citet{DBLP:conf/acl/BansalPG18} leveraged the first characteristic to improve text summarization by performing multi-task learning of summarization with QG, as both these two tasks require the ability to search for salient information in the passage. ~\citet{DBLP:conf/emnlp/DuanTCZ17} applied the second characteristic to improve QA. For an input question $q$ and a candidate answer $\hat a$, they generate a question $\hat q$ for $\hat a$ by way of QG system. Since the generated question $\hat q$ is closely related to $\hat a$, the similarity between $q$ and $\hat q$ helps to evaluate whether $\hat a$ is the correct answer. 

Other works focus on jointly training to combine QG and QA. ~\citet{DBLP:journals/corr/WangYT17} simultaneously train the QG and QA models in the same Seq2Seq model by alternating input data between QA and QG examples. ~\citet{DBLP:conf/naacl/TangDYZSLLZ18} proposed a training algorithm that generalizes Generative Adversarial Network (GANs)~\cite{DBLP:conf/nips/GoodfellowPMXWOCB14} under the question answering scenario. The model improves QG by incorporating an additional QA-specific loss, and improving QA performance by adding artificially generated training instances from QG. However, while joint training has shown some effectiveness, due to the mixed objectives, its performance on QG are lower than the state-of-the-art results, which leaves room for future exploration. 

\subsection{Wider Input Modalities}
\label{sec:input_modalities}

QG work now has incorporated input from knowledge bases (KBQG) and images (VQG). 

Inspired by the use of SQuAD as a question benchmark, ~\citet{DBLP:conf/acl/SerbanGGACCB16} created a 30M large-scale dataset of \textit{(KB triple, question)} pairs to spur KBQG work.  They baselined an attention seq2seq model to generate the target factoid question.  Due to KB sparsity, many entities and predicates are unseen or rarely seen at training time. ~\citet{DBLP:conf/naacl/ElSaharGL18} address these \textit{few-/zero-shot} issues by applying the copying mechanism and incorporating textual contexts to enrich the information for rare entities and relations. 
Since a single KB triple provides only limited information, KB-generated questions also {\it overgeneralize} --- a model asks ``Who was born in New York?" when given the triple \textit{(Donald\_Trump, Place\_of\_birth, New\_York)}. To solve this, ~\citet{DBLP:conf/eacl/KhapraRJR17} enrich the input with a sequence of keywords collected from its related triples. 

Visual Question Generation (VQG) is another emerging topic which aims to ask questions given an 
image. 
We categorize VQG into \textit{grounded-} and \textit{open-ended} VQG by the level of cognition.  
Grounded VQG generates \textit{visually grounded} questions, \textit{i.e.}, all relevant information for the answer can be found in the input image~\cite{DBLP:conf/ijcai/ZhangQYYZ17}. 
A key purpose of grounded VQG is to support the dataset construction for VQA. To ensure the questions are grounded, existing systems rely on image captions to varying degrees. ~\citet{DBLP:conf/nips/RenKZ15} and ~\citet{DBLP:conf/cvpr/ZhuGBF16} simply convert image captions into questions using rule-based methods with textual patterns.
~\citet{DBLP:conf/ijcai/ZhangQYYZ17} proposed a neural model that can generate questions with diverse types for a single image, using separate networks to construct dense image captions and to select question types. 

In contrast to grounded QG, humans ask higher cognitive level questions about what can be inferred rather than what can be seen from an image.  
Motivated by this, ~\citet{DBLP:conf/acl/MostafazadehMDM16} proposed open-ended VQG that aims to generate natural and engaging questions about an image. These are deep questions that require high cognition such as analyzing and creation. 
With significant progress in deep generative models, marked by variational auto-encoders (VAEs) and GANs, such models are also used in open-ended VQG to bring ``creativity'' into generated questions~\cite{DBLP:conf/cvpr/JainZS17,DBLP:conf/coling/FanWWLH18}, showing promising results. This also brings hope to address deep QG from text, as applied in NLG: \textit{e.g.},  SeqGAN~\cite{DBLP:conf/aaai/YuZWY17} and LeakGAN~\cite{DBLP:conf/aaai/GuoLCZYW18}. 

\vspace{-0.1cm}
\subsection{Generation of Deep Questions}
\label{sec:Limitations}

Endowing a QG system with the ability to ask deep questions will help us build curious machines that can interact with humans in a better manner. However, ~\citet{DBLP:conf/cicling/RusCG07} pointed out that asking high-quality deep questions is difficult, even for humans.  Citing the study from~\citet{graesser1994question} to show that students in college asked only about $6$ deep-reasoning questions per hour in a question--encouraging tutoring session. These deep questions are often about events, evaluation, opinions, syntheses or reasons, corresponding to higher-order cognitive levels.

To verify the effectiveness of existing NQG models in generating deep questions, ~\citet{DBLP:conf/icwsm/ChenYHH18} conducted an empirical study that applies the attention Seq2Seq model on LearningQ, a deep-question centric dataset containing over 60$\%$ questions that require reasoning over multiple sentences or external knowledge to answer. 
However, the results were poor; 
the model achieved miniscule BLEU-4 scores of $< 4$ and METEOR scores of $< 9$, compared with $> 12$ (BLEU-4) and $> 16$ (METEOR) on SQuAD. Despite further in-depth analysis are needed to explore the reasons behind, we believe there are two plausible explanations: (1) Seq2Seq models handle long inputs ineffectively, and (2) Seq2Seq models lack the ability to reason over multiple pieces of information. 

Despite still having a long way to go, some works have set out a path forward. A few early QG works attempted to solve this through building deep semantic representations of the entire text, using concept maps over keywords~\cite{DBLP:journals/dad/OlneyGP12} or minimal recursion semantics~\cite{yao2010question} to reason over concepts in the text. ~\citet{DBLP:conf/acl/LabutovBV15} proposed a crowdsourcing-based workflow that involves building an intermediate ontology for the input text, soliciting question templates through crowdsourcing, and generating deep questions based on template retrieval and ranking. Although this process is semi-automatic, it provides a practical and efficient way towards deep QG. 
In a separate line of work, ~\citet{DBLP:conf/nips/RotheLG17} proposed a framework that simulates how people ask deep questions by treating questions as formal programs that execute on the state of the world, outputting an answer. 

Based on our survey, we believe the roadmap towards deep NGQ points towards research that will 
(1) enhance the NGQ model with the ability to consider relationships among multiple source sentences, (2) explicitly model typical reasoning patterns, and (3) understand and simulate the mechanism behind human question asking. 

\vspace{-0.1cm}
\section{Conclusion -- What's the Outlook?}
\label{sec:FutureDirections}
\vspace{-0.1cm}

We have presented a comprehensive survey of NQG, categorizing current NQG models based on different QG-specific and common technical variations, and summarizing three emerging trends in NQG: multi-task learning, wider input modalities, and deep question generation. 

What's next for NGQ?  We end with future potential directions by applying past insights to current NQG models; the ``unknown unknown", promising directions yet explored. 

{\bf When to Ask}: Besides learning what and how to ask, in many real-world applications that question plays an important role, such as automated tutoring and conversational systems, learning when to ask become an important issue. In contrast to general dialog management~\cite{DBLP:journals/jcse/LeeJKLL10}, no research has explored when machine should ask an engaging question in dialog. Modeling question asking as an interactive and dynamic process may become an interesting topic ahead. 

{\bf Personalized QG}: Question asking is quite personalized: people with different characters and knowledge background ask different questions. However, integrating QG with user modeling in dialog management or recommendation system has not yet been explored. Explicitly modeling user state and awareness leads us towards personalized QG, which dovetails deep, end-to-end QG with deep user modeling and pairs the dual of generation--comprehension much in the same vein as in the vision--image generation area. 


\bibliography{acl2019}
\bibliographystyle{acl_natbib}

\end{document}